\documentclass{article}
\usepackage{spconf,amsmath,graphicx,hyperref}
\usepackage{amssymb}

\newcommand{\norm}[1]{\left\lVert#1\right\rVert}

\title{GRAPH NEURAL NETWORKS FOR SAMPLING-INVARIANT EMBEDDINGS OF ORGANIZED SIGNAL SETS}
\name{Martin Bauw$^{\star}$ \qquad Santiago Velasco-Forero$^{\dagger}$ \qquad Jesus Gustavo Angulo$^{\ddagger}$\thanks{Corresponding author: martin.bauw@onera.fr.}}
\address{$^{\star}$ DEMR, ONERA, Université Paris-Saclay, 91120 Palaiseau, France \\
$^{\dagger}$ Centre for Statistics and Images (STIM), Mines Paris, PSL University, Fontainebleau, France \\
$^{\ddagger}$ Centre for Applied Mathematics (CMA), Mines Paris, PSL University, Sophia Antipolis, France}
\begin{document}
%
\maketitle
\begin{abstract}
Sensor networks and radars can deliver signals as organized sets, e.g. ordered signals, signals describing range cells within a grid or signals perceived as graph nodes. Within such sets, individual signals may be characterized by distinct sampling parameters. This paper investigates organized signal sets neural network encoders. In the context of this work, the purpose of such encoders is to project heterogeneously sampled signal sets into an arbitrary fixed-size vectors space. This new representation space is designed so that signal sets can be processed as vectors rid of sampling differences to allow for arbitrary topology-aware processing with no signal processing constraints. Within this representation space designed to reduce the influence of heterogeneous sampling parameters, the relevance of signal sets representations is evaluated by considering signal sets discrimination potential with a focus on waveforms separation. The encoding and embeddings discrimination experiments conducted rely exclusively on synthetic complex-valued radiofrequency signals.
\end{abstract}
\section{INTRODUCTION}
\label{sec:intro}

This paper compares encoding methods producing fixed-size vector representations for organized signal sets. The organized signal sets may represent signals produced by sensor networks, signals contained over a radar cells grid, or any graph whose nodes are defined by signals. Encoding a graph of individual signals is particularly interesting in a radar context as it allows for adaptive radar cells processing. This relates to range-Doppler processing, where the content of a neighborhood of range-Doppler cells can be used to detect and discriminate targets. Replacing CFAR detectors which rely on limited context with neural networks that adaptively process radar echoes in larger contexts is an active research question \cite{wang2019study} \cite{bauw2026detecting}.

The fixed-size vector representations motivating this paper aim at transposing a classification or anomaly detection (AD) problem from a signal processing representation space to a data processing representation space in which no sampling parameters heterogeneity hinders the signal sets discrimination task. The encoder this work aims at building is therefore expected to produce sampling-invariant, waveform-aware embeddings, this while remaining sensitive to waveform ordering within sets. This waveforms order awareness led the experiments to include graphs. Related works include \cite{bauw:tel-04106703} which proposed heterogeneously sampled signals encoding for downstream sampling-agnostic discrimination in a radar context. In \cite{liu2021self}, the authors propose training an encoder with a contrastive learning objective to encode and subsequently classify individual signals. Creating generic embeddings for radiofrequency (RF) signal recognition has also been investigated in \cite{henneketowards}, which considers both raw 1D IQs and 2D time-frequency input representations. Both input types have been considered in the literature \cite{o2018over} \cite{brooks2019complex} \cite{scholl2025end} \cite{mazouz2026multi}, the 2D time-frequency input making it easy to leverage image processing-inspired pipelines. The originality of this work stems from the combination of the following elements: the signals being processed as an ordered set and encoded with a specific graph, the signals intra and inter-sets sampling heterogeneity, the choice of raw IQs as input, and the radiofrequency waveforms selected.

\section{SIGNAL SETS ENCODING FORMALIZATION}
\label{sec:formal}

\begin{figure}[htb]
\centering
\includegraphics[width=0.9\columnwidth]{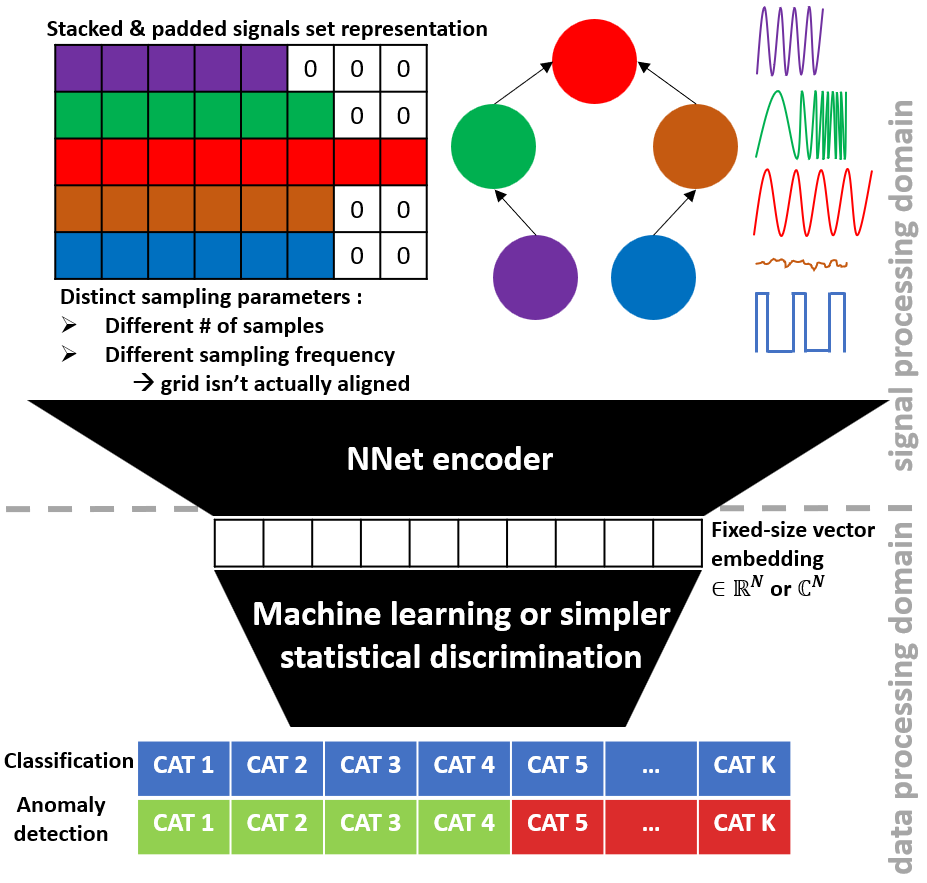}
\caption{The encoding presented processes sets of signals with different sampling parameters. Individual signals within sets are organized as graphs}
\label{fig:sigsgraph2vec}
\end{figure}

A graph neural network (GNN) \cite{scarselli2008graph} $\Phi$ will take the signal set composed of $P$ input $M$-dimensional signals $Z \in \mathbb{C}^{P \times M}$ along with an adjacency matrix defining the graph topology $A^{P \times P}$  (where $A[i,j]$ equals $1$ if the $i$-th signal shares an edge with the $j$-th signal) as input to produce a real-valued latent space representation also called signals set embedding $E \in \mathbb{R}^{N}$. In this setup, input signals carry at most $M$ samples with varying sampling frequencies, and end up represented by latent vectors of size $N$. Individual signals within signal sets have an actual length $M^{'}$ such that $M^{'}\leq M$, $M$ defining a zero-padding length where needed. Zero-padding ensures uniform input dimensions for signals with varying lengths. This setup is illustrated for a set of five signals on Fig. \ref{fig:sigsgraph2vec}. The encoding, which amounts to an inference, of a batch tensor $G$ of $B$ signal sets thus corresponds to the following equation:

\begin{equation}
\label{eq:batch-encoding}
\Phi(G,A) = E \in \mathbb{R}^{B \times N}
\end{equation}

with $G \in \mathbb{C}^{B \times P \times M}$. The adjacency matrix $A$ is required for graph neural network operations: it specifies which neighboring nodes' features are aggregated to update the node representations in the next layer \cite{kipf2017semisupervised}. The aggregation of neighboring nodes features amounts here to an aggregation of signals (e.g. radar cells in a grid). To ensure similar signal sets end up similarly encoded and distinct sets correspond to easily separable embeddings, the loss used to train the signal sets encoder is the Normalized Temperature-Scaled Cross-Entropy loss (NT-Xent) \cite{pmlr-v119-chen20j}. This loss is computed thanks to positive pairs of signal sets $(e_i, e_j)$, i.e. two signal sets sharing the same label:
\begin{equation}
\label{eq:NT-Xent}
\ell(i, j) = -\log \frac{\exp\left(\frac{\text{sim}(e_i, e_j)}{\tau}\right)}{\sum_{k=1}^{B} \mathbf{1}_{[k \neq i]} \exp\left(\frac{\text{sim}(e_i, e_k)}{\tau}\right)}
\end{equation}

In the previous expression, $\tau$ is the temperature parameter and $sim(e_i,e_j) = \frac{e_i^T e_j}{\norm{e_i} \norm{e_j}}$ the cosine similarity. The denominator includes both positive and negative pairs. The NT-Xent loss is a contrastive learning loss pulling the embeddings of pairs sharing labels together while pulling apart pairs of representations corresponding to distinct labels.

\section{SIGNAL SETS ENCODING IMPLEMENTATION AND EVALUATION}
\label{sec:implem}

Organized signal sets were represented with and without a neural network encoder to produce a comparison revealing the relevance of neural networks and graphs for the discrimination task considered. The following section details which representations and encoding approaches were investigated. The downstream discrimination evaluating the embeddings relevance is a signal sets templates (cf. Table \ref{tab:templates}) classification.

\subsection{Graph neural network implementation}

The main motivation of this work being the production of discriminative representations of organized signal sets, graph neural networks \cite{scarselli2008graph} appeared as the most natural deep learning approach. The proposed encoder is a graph convolutional network (GCN) \cite{kipf2017semisupervised} trained with the contrastive learning objective \eqref{eq:NT-Xent}. The GCN is implemented with three layers and produces a 128-dim signals set embedding thanks to mean-pooling. This architecture carries 2.33M parameters and can be said to be a graph2vec \cite{narayanan2017graph2vec} \cite{grohe2020word2vec} encoding. Inspired by radar test cell-centric processing, the graph topology eventually selected as the method put forward is a chain-like graph with directed edges oriented towards the central node, as illustrated on Fig. \ref{fig:sigsgraph2vec}.

\subsection{Neural network baselines}

Several deep learning baselines were considered in our initial experiments. A graph autoencoder (GAE) \cite{kipf2016variational} \cite{pmlr-v97-gao19a} leveraging the topology defined by the adjacency matrix and a regular autoencoder (AE) ignoring the latter were trained along with the proposed contrastive GCN. The graph autoencoder implemented failed to achieve any substantial signal sets separability and thus will be absent from the presented results. Evaluating regular autoencoders using stacked, zero-padded signals as 2D inputs aims at including a non-graph neural network encoder to our baselines. The 1D AE eventually selected relies on 1D convolutions spanning the temporal axis of individual signals within a signals set, and carries 1.28M parameters. It processes the five complex-valued signals of a set as 10 inputs channels through a real and imaginary parts separation. Training is  conducted with a length-masked mean squared error (MSE) reconstruction loss, and the signal sets representations are extracted from the 128-dim bottleneck. The order of the set remains available to the AE through the signals stacking order leading to the complex-valued 2D input representation.

\subsection{Features engineering baselines}

Machine learning-free representations were defined to evaluate the actual contribution of approaches relying on a trained neural network to produce a representation with discriminative power. The two retained baselines rely on spectral (baseline name \textit{raw\_spectrum}) and statistical signal (baseline name \textit{raw\_dsp}) features \cite{nandi1998algorithms}. The spectral baseline extracts 320 features from Fourier bins computed over individual signals within sets, whereas the statistical signal baseline combines temporal and spectral representations attributes to produce 44 features as a signals set representation.

\subsection{Evaluating the encoding quality without classification/AD}

\begin{figure}[htb]
\centering
\includegraphics[width=0.8\columnwidth]{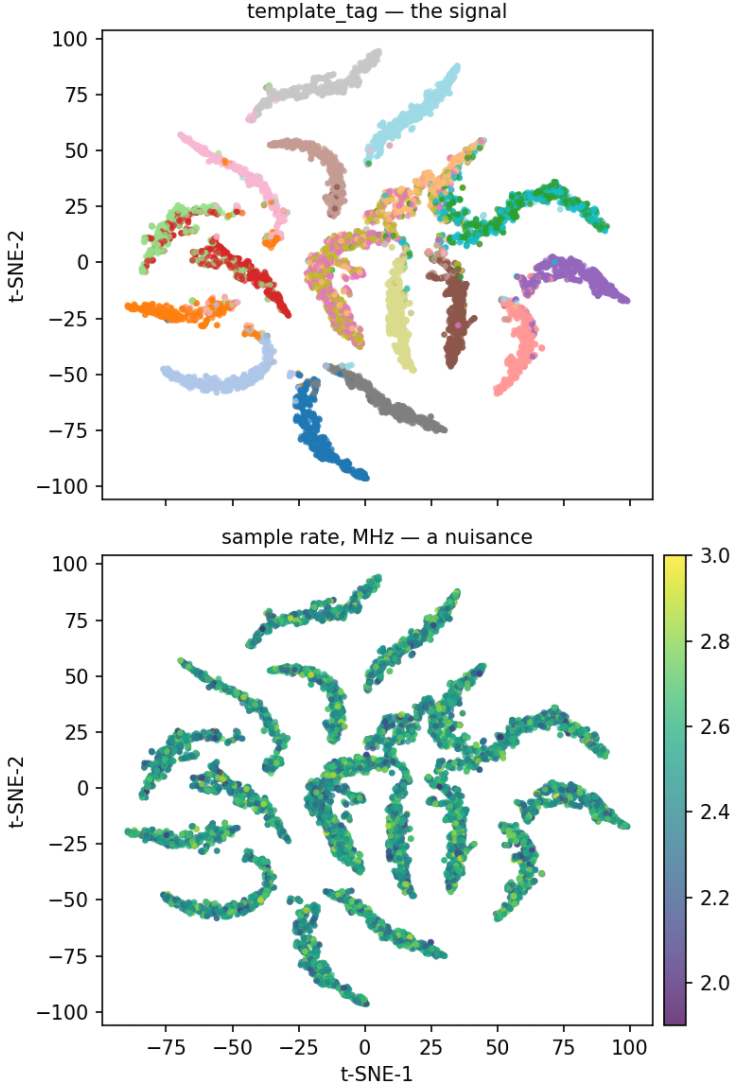}
\caption{Latent space 2D distribution of test signal sets, computed with t-SNE. \textbf{Top:} colormap describes organized signal sets templates that should be separable. \textbf{Bottom:} colormap describes a signal sampling parameter that should not impact the latent space distribution.}
\label{fig:latent}
\end{figure}

In the case of neural network encoders, the silhouette score \cite{rousseeuw1987silhouettes} and k-nearest neighbors (KNN) were used to quantitatively evaluate the relevance of the latent space distribution before classification. 
%
%
The training signal sets templates separability was thus evaluated with two methods relying on Euclidean distances in the representation space. These two scores motivated the choice of both the GCN and the AE encoders among the different neural networks of each type trained. The separability of distinct signal sets templates could additionally be qualitatively evaluated through 2D latent space distribution visualization thanks to PCA, t-SNE and UMAP-provided dimensionality reductions. One such latent space visualization is depicted on Fig. \ref{fig:latent}.

\section{SYNTHETIC SIGNAL SETS GENERATION}
\label{sec:data}

The signal sets used for training and evaluating the proposed approaches are synthetic, i.e. they were generated from controlled parameters. Each signals set includes five signals, each with possibly unshared sampling parameters within the set. Each set corresponds to a multiple waveforms template, all of which are defined in Table \ref{tab:templates}. A positive pair in \eqref{eq:NT-Xent} is defined by a common template. Each of the 19 templates fixes an ordered five-node sequence, and node order is preserved at generation time so the graph depicted on Fig. \ref{fig:sigsgraph2vec} reads the set as a line with a centre.

The individual signals generation parameters, presented in \ref{tab:datasets}, are selected to ensure training, validation and test sets are mutually exclusive with respect to sampling frequencies, signal lengths and SNR (Signal-to-Noise Ratio). Each individual signal within a set independently draws its length $M^{'}$ and sampling rate $f_s$ from the dataset valid values pool, the SNR is drawn once per set and shared by its five nodes. Valid signal parameters pools are regular grids with the
held-out values excised, so Train$^{+}$, Val and Test are pairwise mutually exclusive for all three parameters. Train$^{+}$ is an enriched training set with increased sampling and noise parameters created to investigate the performances improvements allowed by increased training diversity while keeping the same number of training signal sets. Train and Train$^{+}$ are thus alternative training sets and are never used together.

\begin{table}[t]
\caption{Signal set templates. Abbreviations: C$\uparrow$/C$\downarrow$ linear up/down chirp, Q$\uparrow$/Q$\downarrow$ quadratic up/down chirp, $\eta$ complex Gaussian noise.}
\label{tab:templates}
\centering
\footnotesize
\begin{tabular}{@{}ll@{}}
\hline
Class & Node sequence \\
\hline
\texttt{cw}                 & CW CW CW CW CW \\
\texttt{am}                 & AM AM AM AM AM \\
\texttt{fsk}                & FSK FSK FSK FSK FSK \\
\texttt{bpsk}               & BPSK BPSK BPSK BPSK BPSK \\
\texttt{chirp\_up}          & C$\uparrow$ C$\uparrow$ C$\uparrow$ C$\uparrow$ C$\uparrow$ \\
\texttt{chirp\_down}        & C$\downarrow$ C$\downarrow$ C$\downarrow$ C$\downarrow$ C$\downarrow$ \\
\texttt{chirp\_quad\_up}    & Q$\uparrow$ Q$\uparrow$ Q$\uparrow$ Q$\uparrow$ Q$\uparrow$ \\
\texttt{chirp\_quad\_down}  & Q$\downarrow$ Q$\downarrow$ Q$\downarrow$ Q$\downarrow$ Q$\downarrow$ \\
\noalign{\smallskip}
\texttt{chirp3\_fsk2}       & C$\uparrow$ C$\uparrow$ C$\uparrow$ FSK FSK \\
\texttt{fsk3\_chirpup2}     & FSK FSK FSK C$\uparrow$ C$\uparrow$ \\
\texttt{chirpup2\_fsk3}     & C$\uparrow$ C$\uparrow$ FSK FSK FSK \\
\texttt{chirp3\_noise2}     & C$\uparrow$ C$\uparrow$ C$\uparrow$ $\eta$ $\eta$ \\
\texttt{noise2\_chirp3}     & $\eta$ $\eta$ C$\uparrow$ C$\uparrow$ C$\uparrow$ \\
\texttt{noise\_chirp\_noise}& $\eta$ C$\uparrow$ C$\uparrow$ C$\uparrow$ $\eta$ \\
\texttt{am\_fsk\_am}        & AM FSK AM FSK AM \\
\texttt{chirp\_up\_down}    & C$\uparrow$ C$\downarrow$ C$\uparrow$ C$\downarrow$ C$\uparrow$ \\
\texttt{chirp\_up\_down\_quad} & Q$\uparrow$ Q$\downarrow$ Q$\uparrow$ Q$\downarrow$ Q$\uparrow$ \\
\noalign{\smallskip}
\texttt{cw\_chirp\_cw}      & AM C$\uparrow$ C$\downarrow$ C$\uparrow$ AM \\
\texttt{mixed}              & C$\uparrow$ $\eta$ FSK CW AM \\
\hline
\end{tabular}
\end{table}

\begin{table}[t]
\caption{Dataset generation parameters. All signals are
complex baseband, corrupted by additive white Gaussian noise, and
zero-padded to $M=4096$ samples.}
\label{tab:datasets}
\centering
\footnotesize
\begin{tabular}{@{}lrrrr@{}}
\hline
 & Train & Train$^{+}$ & Val & Test \\
\hline
Sets                & 100\,000 & 100\,000 & 10\,000 & 10\,000 \\
Seed                & 4242 & 5309 & 7131 & 9468 \\
\noalign{\smallskip}
$N$ (samples)       & 2048--4096 & 1920--4096 & 2150--4030 & 2304--4007 \\
\quad pool size     & 13 & 65 & 4 & 4 \\
$f_s$ (MHz)         & 1.70--3.10 & 1.60--3.20 & 1.95--3.05 & 1.90--3.00 \\
\quad pool size     & 11 & 57 & 4 & 4 \\
SNR (dB)            & 7--30 & 5--32 & 10--29 & 9--28 \\
\quad pool size     & 4 & 47 & 4 & 4 \\
\hline
\end{tabular}
\end{table}

\section{EXPERIMENTS AND RESULTS}
\label{sec:results}

\begin{figure}[htb]
\centering
\includegraphics[width=\columnwidth]{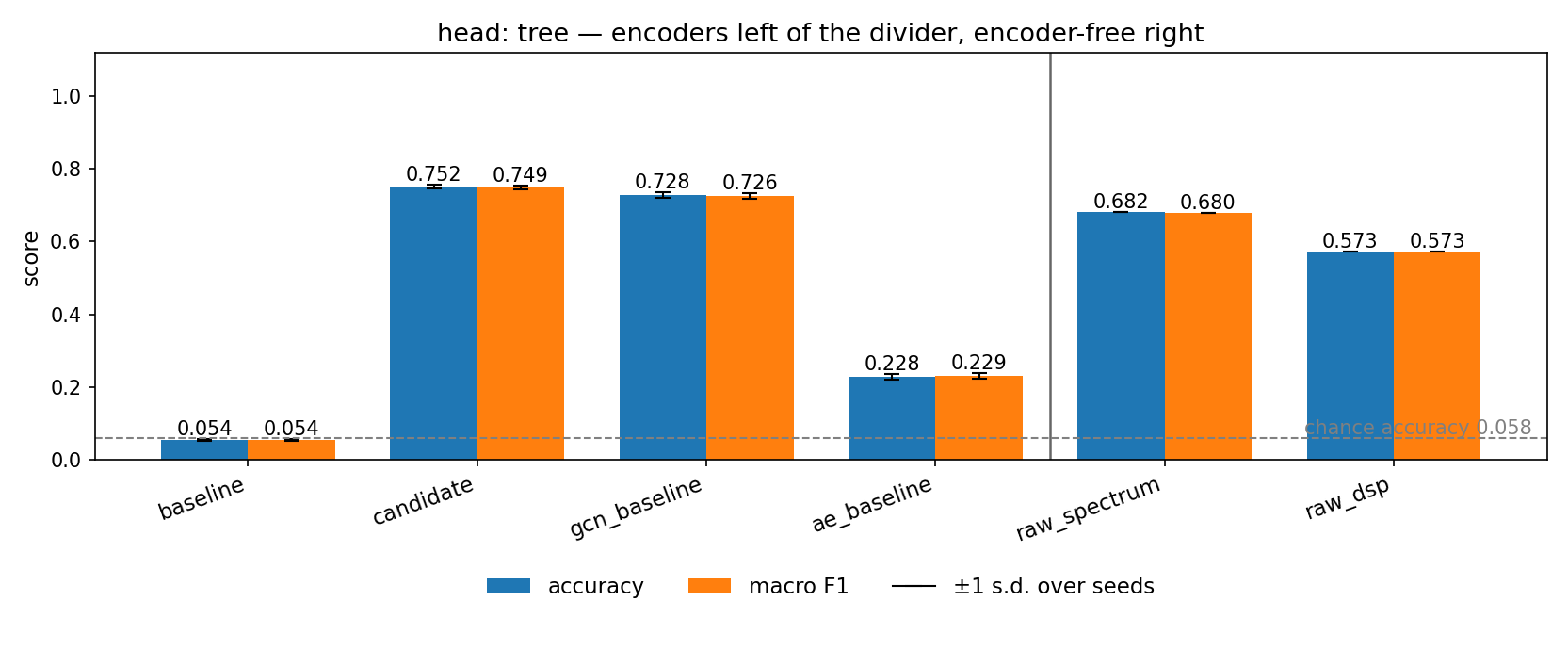}
\caption{Performance of a decision tree classifier on embeddings from: GCN candidate, GCN baseline, AE (1D autoencoder), raw spectrum (spectral features), and raw dsp (statistical features). Neural networks metrics are averaged over three seeds and the error bar depicts $\pm$ std.}
\label{fig:metrics}
\end{figure}

The proposed GCN encoder and the selected baselines lead to the performances presented in Fig. \ref{fig:metrics}. These performances correspond to a decision tree, one of the classification heads considered for latent space classification. The GCN encoder trained on the Train$^{+}$ dataset over 60 epochs leads to the best signal sets templates classification performances in the latent representation space, closely followed by the GCN baseline trained on the Train dataset (cf. Table \ref{tab:datasets}), while the AE trained over 40 epochs struggles and ranks below spectral and statistical signal features. The poor AE performances are not surprising as the training objective is generative and not contrastive for this baseline. The reconstruction objective fails at producing a competitive representation. The comparison between the AE baseline and the GCN encoders amounts to comparing a general-purpose encoder with encoders specifically trained for the discrimination implemented by the downstream classification task. For the three neural network encoders, the validation set was used to adjust training hyperparameters and avoid using an overfitted encoder to produce signal sets embeddings for the test set.

\begin{figure}[htb]
\centering
\includegraphics[width=0.85\columnwidth]{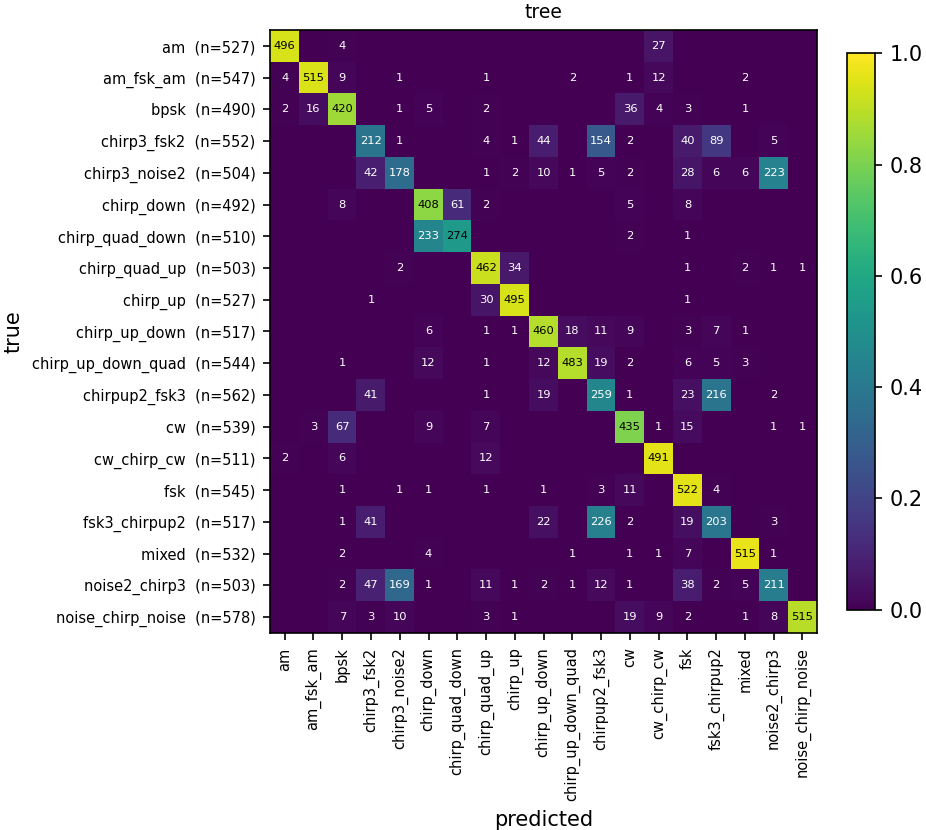}
\caption{Confusion matrix for the latent space classification task. The performances reported stem from a classification head implemented with a decision tree.}
\label{fig:confusion}
\end{figure}

The test set confusion matrix associated with the decision tree classification head on top of the trained GCN encoder is depicted on Fig. \ref{fig:confusion}. It reveals a limited confusion between linear (template \textit{chirp\_down}) and quadratic (template \textit{chirp\_quad\_down}) chirps with decreasing frequencies. More interestingly, it shows the best and proposed approach still struggles to identify distinct waveforms orders within the signal sets : \textit{chirp3\_noise2} is substantially confused with \textit{noise2\_chirp3}, and a similar confusion can be observed for \textit{chirpup2\_fsk3} and \textit{fsk3\_chirpup2}. In the experiments conducted, the \textit{raw\_spectrum} and \textit{raw\_dsp} baselines were unexpectedly hard to beat, and in some cases with different signal sets templates not presented here, systematically performed better in downstream classification with respect to trained neural encoders.

\section{CONCLUSION}
\label{sec:conclusion}

This work put forward the relevance of graph neural networks to encode organized signal sets for downstream discrimination. The proposed encoder produces discriminative representations that exhibit robustness to heterogeneous sampling parameters under the synthetic conditions investigated. Taking into account a graph structure allows for the representations to be order-sensitive within the signals set, making the presented method relevant for sensor networks and radar cells neighborhoods processing. The experiments conducted additionally reaffirmed the relative competitiveness of spectral and statistical signal features as embeddings for the classification task studied. Future work will consider larger and more complex signal sets, will include real data and more diverse waveforms, and will evaluate the relevance of the produced embeddings with both classification and anomaly detection as the latent space discrimination tasks. The generalization of latent space representations to waveforms unseen during training remains challenging, suggesting future work on graph topologies and attention mechanisms.

%


\clearpage
\bibliographystyle{IEEEbib}
\bibliography{ICASSP2027_mbauw_submission_refs}

\begin{thebibliography}{10}

\bibitem{wang2019study}
Li~Wang, Jun Tang, and Qingmin Liao,
\newblock ``A study on radar target detection based on deep neural networks,''
\newblock {\em IEEE Sensors Letters}, vol. 3, no. 3, pp. 1--4, 2019.

\bibitem{bauw2026detecting}
Martin Bauw,
\newblock ``Detecting radar targets swarms in range profiles with a partially
  complex-valued neural network,''
\newblock in {\em 2026 34th European Signal Processing Conference (EUSIPCO)}.
  IEEE, 2026.

\bibitem{bauw:tel-04106703}
Martin Bauw,
\newblock {\em {One-class classification for low resolution targets
  discrimination with limited supervision in pulse Doppler radars}},
\newblock Theses, {Universit{\'e} Paris sciences et lettres}, Jan. 2023.

\bibitem{liu2021self}
Dongxin Liu, Peng Wang, Tianshi Wang, and Tarek Abdelzaher,
\newblock ``Self-contrastive learning based semi-supervised radio modulation
  classification,''
\newblock in {\em Milcom 2021-2021 IEEE military communications conference
  (milcom)}. IEEE, 2021, pp. 777--782.

\bibitem{henneketowards}
Lukas Henneke and Frank Kurth,
\newblock ``Towards generic embeddings for robust rf signal recognition,''
\newblock in {\em 2026 34th European Signal Processing Conference (EUSIPCO)}.
  IEEE, 2026.

\bibitem{o2018over}
Timothy~James O’Shea, Tamoghna Roy, and T~Charles Clancy,
\newblock ``Over-the-air deep learning based radio signal classification,''
\newblock {\em IEEE Journal of Selected Topics in Signal Processing}, vol. 12,
  no. 1, pp. 168--179, 2018.

\bibitem{brooks2019complex}
Daniel~A Brooks, Olivier Schwander, Fr{\'e}d{\'e}ric Barbaresco, Jean-Yves
  Schneider, and Matthieu Cord,
\newblock ``Complex-valued neural networks for fully-temporal micro-doppler
  classification,''
\newblock in {\em 2019 20th International Radar Symposium (IRS)}. IEEE, 2019,
  pp. 1--10.

\bibitem{scholl2025end}
Stefan Scholl, Chandana Panati, and Simon Wagner,
\newblock ``End-to-end learning for radar electronic support: Multilabel
  classification and explainable ai,''
\newblock {\em IEEE Transactions on Aerospace and Electronic Systems}, vol. 61,
  no. 5, pp. 11128--11140, 2025.

\bibitem{mazouz2026multi}
Reihan Mazouz, Jos{\'e} Picheral, Abigael Taylor, Jonathan Bosse, Sylvie
  Marcos, and Chakib Belafdil,
\newblock ``Multi-resolution spectrograms detection of lpi radar with
  time-frequency attention augmented yolo,''
\newblock in {\em ICASSP 2026-2026 IEEE International Conference on Acoustics,
  Speech and Signal Processing (ICASSP)}. IEEE, 2026, pp. 5926--5930.

\bibitem{scarselli2008graph}
Franco Scarselli, Marco Gori, Ah~Chung Tsoi, Markus Hagenbuchner, and Gabriele
  Monfardini,
\newblock ``The graph neural network model,''
\newblock {\em IEEE transactions on neural networks}, vol. 20, no. 1, pp.
  61--80, 2008.

\bibitem{kipf2017semisupervised}
Thomas~N. Kipf and Max Welling,
\newblock ``Semi-supervised classification with graph convolutional networks,''
\newblock in {\em International Conference on Learning Representations}, 2017.

\bibitem{pmlr-v119-chen20j}
Ting Chen, Simon Kornblith, Mohammad Norouzi, and Geoffrey Hinton,
\newblock ``A simple framework for contrastive learning of visual
  representations,''
\newblock in {\em Proceedings of the 37th International Conference on Machine
  Learning}. 13--18 Jul 2020, vol. 119 of {\em Proceedings of Machine Learning
  Research}, pp. 1597--1607, PMLR.

\bibitem{narayanan2017graph2vec}
Annamalai Narayanan, Mahinthan Chandramohan, Rajasekar Venkatesan, Lihui Chen,
  Yang Liu, and Shantanu Jaiswal,
\newblock ``graph2vec: Learning distributed representations of graphs,''
\newblock {\em arXiv preprint arXiv:1707.05005}, 2017.

\bibitem{grohe2020word2vec}
Martin Grohe,
\newblock ``word2vec, node2vec, graph2vec, x2vec: Towards a theory of vector
  embeddings of structured data,''
\newblock in {\em proceedings of the 39th ACM SIGMOD-SIGACT-SIGAI symposium on
  principles of database systems}, 2020, pp. 1--16.

\bibitem{kipf2016variational}
Thomas~N. Kipf and Max Welling,
\newblock ``Variational graph auto-encoders,''
\newblock {\em arXiv preprint arXiv:1611.07308}, 2016.

\bibitem{pmlr-v97-gao19a}
Hongyang Gao and Shuiwang Ji,
\newblock ``Graph u-nets,''
\newblock in {\em Proceedings of the 36th International Conference on Machine
  Learning}, Kamalika Chaudhuri and Ruslan Salakhutdinov, Eds. 09--15 Jun 2019,
  vol.~97 of {\em Proceedings of Machine Learning Research}, pp. 2083--2092,
  PMLR.

\bibitem{nandi1998algorithms}
Asoke~K Nandi and Elsayed~Elsayed Azzouz,
\newblock ``Algorithms for automatic modulation recognition of communication
  signals,''
\newblock {\em IEEE Transactions on communications}, vol. 46, no. 4, pp.
  431--436, 1998.

\bibitem{rousseeuw1987silhouettes}
Peter~J Rousseeuw,
\newblock ``Silhouettes: a graphical aid to the interpretation and validation
  of cluster analysis,''
\newblock {\em Journal of computational and applied mathematics}, vol. 20, pp.
  53--65, 1987.

\end{thebibliography}

\end{document}